\documentclass[runningheads]{llncs}
\usepackage{marvosym}
 
\usepackage{eccv}

\usepackage{eccvabbrv}
\usepackage{multicol}
\usepackage{multirow}
\usepackage{graphicx}
\usepackage{booktabs}
\usepackage[accsupp]{axessibility}  
\usepackage{algorithm}
\usepackage{algorithmic}
\usepackage{subcaption}  

\usepackage{hyperref}

\usepackage{orcidlink}
\usepackage{enumitem}

\begin{document}

\title{Calibrated Harmonic Overlaid Implicit Neural Representations for Multi-Dimensional Data} 

\titlerunning{CHOIR for Multi-Dimensional Data}

\author{Honghang Chen\inst{1}\orcidlink{0009-0009-5885-2074} \and
Xiujun Zhang\inst{2}\textsuperscript{(\text{\Letter})}\orcidlink{0009-0007-4758-4727} \and
Xiaoli Sun\inst{1}\textsuperscript{(\text{\Letter})}\orcidlink{0000-0002-3462-4343} \and Mingqing~Xiao\inst{3}\orcidlink{0000-0003-3241-4112}}

\authorrunning{H.~Chen et al.}

\institute{Shenzhen University, Shenzhen, China
\and
Shenzhen Polytechnic University, Shenzhen, China 
\and
Southern Illinois University Carbondale, USA\\
\email{honghangchenhaha@gmail.com}
}


\maketitle
\begingroup
\renewcommand{\thefootnote}{\Letter}
\footnotetext{Corresponding author.}
\endgroup

\begin{abstract}
Implicit neural representation (INR) has emerged as a powerful prior for multi-dimensional data (e.g., multispectral images and videos). However, most INR methods employing periodic activation functions (e.g., Sine) predominantly rely on function composition. This mechanism introduces optimization instability as network depth increases, thereby limiting their performance. Meanwhile, these methods fail to incorporate proper physical priors to effectively alleviate spectrum bias. To address these issues, inspired by the commonalities between deep periodic networks and generalized Fourier series, we propose a novel Calibrated Harmonic Overlaid Implicit Neural Representation (CHOIR). Specifically, we utilize Coordinated Harmonic Superposition (CHS) to replace the conventional function composition used in most INRs, thereby ensuring optimization stability when scaling network depth. Furthermore, we introduce a Perceptual Spectrum Calibration (PSC) to mitigate spectrum bias. This calibration embeds the ubiquitous power-law spectrum prior of natural images and adjusts the globally fixed spectrum towards a physically plausible log-uniform distribution. Extensive experiments on various multidimensional data recovery problems demonstrate that our method achieves superior performance over state-of-the-art approaches. Code is available at \url{https://github.com/chorl0229/CHOIR}.

\end{abstract}

\section{Introduction}

Implicit Neural Representations (INR) have become a powerful paradigm for multi-dimensional data recovery, such as multispectral images, videos, and other forms of data, by parameterizing the signal as a continuous function of coordinates~\cite{sitzmann2020implicit,saragadam2023wire,liu2024finer,li2024superpixel,zhou2025frequency,rezaeian2025sl2a}. Among them, methods employing periodic activation functions (e.g., sine functions)~\cite{sitzmann2020implicit} enable efficient modeling of high-frequency components and exhibit prominent advantages in representing high-frequency details, which has made them one of the important research directions in this field. Although subsequent works have continuously improved them through position encoding~\cite{tancik2020fourier,mildenhall2021nerf,kania2025fresh} or activation function design~\cite{saragadam2023wire,liu2024finer,rezaeian2025sl2a}, the deep scalability of INR has not been fully explored, and spectral bias calibration matched to natural-image statistics remains underexplored.

\begin{figure}[htbp]
  \centering
  \includegraphics[width=\linewidth]{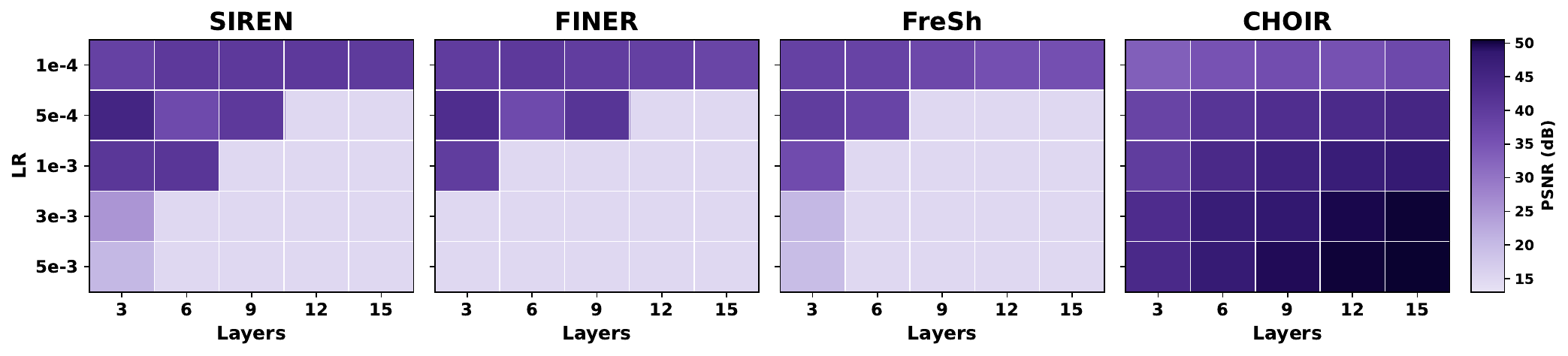}
  \caption{PSNR heatmap vs. network depth and learning rate for sine-based INR methods under data missing completion (random missing and observation rate OR=0.30) on MSI Flowers dataset.}
  \label{fig:heatmap}
\end{figure}

In response, we reveal the reasons for the limitations of these methods from the perspective of harmonic signal representation. First, most periodic INR methods generally adopt a layer-wise function composition $\Phi_\theta = \phi_L \circ \cdots \circ \phi_1$, whereas the superposition principle of signals requires the components to be additive. This architectural mismatch leads to a bottleneck: gradient pathologies that destabilize deep network optimization and constrain model expressiveness. Furthermore, the energy distribution of natural signals generally follows a $1/f$ power-law statistic~\cite{field1987relations}, while most INR methods constrain all neurons with a globally fixed frequency scaling $\omega_0$. This gives rise to another bottleneck: unconstrained frequency distributions that impede model convergence. Resolving these critical limitations to enable stable and scalable deep periodic INRs constitutes our primary contribution. Figure~\ref{fig:heatmap} effectively confirms this problem. As shown, the PSNR of our method increases gradually with the network depth and achieves the highest value at 15 layers. All subsequent experiments use 12 layers for performance-efficiency trade-off. In contrast, other methods, such as SIREN~\cite{sitzmann2020implicit}, FINER~\cite{liu2024finer} and FreSh~\cite{kania2025fresh}, reach their peak PSNR at relatively shallow layers. A deeper network brings no performance gain for these methods; instead, it causes a clear performance degradation. Meanwhile, our method exhibits stronger robustness to the learning rate than other methods.

Inspired by the inherent equivalence between deep periodic INR and generalized Fourier series, we replace the function composition paradigm with hybrid composition-superposition architecture and propose \textbf{Calibrated Harmonic Overlaid Implicit Neural Representations (CHOIR)}. CHOIR includes two key mechanisms: Coordinated Harmonic Superposition (CHS), which eliminates optimization instability caused by deep networks via implicit curriculum learning, and Perceptual Spectrum Calibration (PSC), which embeds the $1/f$ power-law spectrum prior into frequency allocation to systematically alleviate spectrum bias. The core contributions of this paper are summarized as follows:
\begin{enumerate}[label=(\roman*)]
    \item We theoretically reveal the fundamental contradiction between the function composition architecture of deep periodic INR and the superposition principle of signals, clarifying the essential reason for the deep scalability limitation of most INR methods.
    \item We propose the Coordinated Harmonic Superposition (CHS) mechanism, which enables stable deep scaling of hybrid composition-superposition architecture through implicit curriculum learning, fully unleashing the representational potential of deep periodic INR.
    \item We introduce the Perceptual Spectrum Calibration (PSC) strategy, embedding the $1/f$ power-law spectrum prior as a physical calibration into frequency learning, significantly improving high-frequency detail reconstruction in multi-dimensional data recovery tasks.
\end{enumerate}

\section{Related Work}

\subsection{Implicit Neural Representations}
INR parameterizes signals as continuous coordinate mappings, emerging as a powerful paradigm for multi-dimensional data representation. However, frequency principles \cite{xu2019frequency, rahaman2019spectral} reveal that ReLU-based MLPs inherently prioritize low-frequency components during training, introducing a bias against high-frequency signals. To alleviate this limitation, existing research has primarily focused on two directions: advances in position encoding and advances in activation functions.

On the position encoding front, Gaussian Fourier feature encoding \cite{tancik2020fourier} lifts low-dimensional coordinates into a high-dimensional frequency space via random mappings. NeRF \cite{mildenhall2021nerf} popularized sine position encoding for novel view synthesis, laying the foundation for modern coordinate encoding schemes. Subsequent works further refined this line of research. BACON \cite{lindell2022bacon} proposed band-limited coordinate networks to control spectrum bandwidth for multi-scale representation. PNF~\cite{yang2022polynomial} enables controllable subband decomposition and manipulation, while BANF~\cite{shabanov2024banf} realizes band-limited filtering for level-of-detail reconstruction. Instant-NGP \cite{muller2022instant} introduced multi-resolution hash encoding to drastically accelerate training convergence. FreSh \cite{kania2025fresh} adaptively aligned frequency initialization with input data. Additionally, batch normalization has been shown to effectively mitigate spectrum bias in coordinate networks \cite{cai2024batch}.

Regarding activation functions, SIREN \cite{sitzmann2020implicit} pioneered the use of sine activations for efficient high-frequency representation, establishing the theoretical framework for periodic INR. MSIREN~\cite{mehta2021modulated} further modulates periodic activations for generalizable local representations, and INCODE~\cite{kazerouni2024incode} conditions sinusoidal activations with prior-knowledge embeddings. WIRE \cite{saragadam2023wire} extended this by employing continuous Gabor wavelet activations to jointly model time-frequency domains, while \cite{ramasinghe2022beyond} constructed a unified framework that transcends pure periodicity using Gaussian activations. MIRE~\cite{jayasundara2025mire} matches layer-wise activation functions to target signals through dictionary learning. Recent efforts have focused on adaptive and constrained designs. MFN~\cite{fathony2021multiplicative} applies multiplicative filters directly to the input coordinates and admits a basis-expansion interpretation. However, it still relies on layer-wise nested multiplicative modulation rather than explicitly altering the functional representation through additive superposition. FINER \cite{liu2024finer} introduced variable-period activations for flexible spectrum adjustment. STAF~\cite{morsali2025staf} parameterized activations as trainable Fourier-series functions for adaptive spectral learning. SASNet~\cite{feng2026sasnet} localized sinusoidal responses with spatially adaptive masks to reduce frequency leakage. \cite{shi2024improved} proposed Fourier reparameterization to impose structural constraints on weights in the frequency domain. SL2A-INR \cite{rezaeian2025sl2a} leveraged a single-layer learnable activation for adaptive frequency response, and IGA-INR \cite{shi2025inductive} further reduced spectrum bias through inductive gradient adjustments.

\subsection{INR-based Methods in Multi-dimensional Data}

Using the network structure itself as an implicit prior has emerged as an important research direction in multi-dimensional data recovery. Deep Image Prior (DIP)~\cite{ulyanov2018deep} first demonstrated the implicit regularization effect derived from the structure of untrained convolutional networks. CoIL~\cite{sun2021coil} further introduced coordinate-based INR into imaging inverse problems, revealing the potential of MLP-based continuous coordinate mappings for image recovery tasks.
Inspired by these advances, treating INR as a continuous prior has become a vital paradigm for multi-dimensional data recovery~\cite{luo2025continuous}. In this paradigm, discrete factor matrices are replaced by MLP-parameterized continuous factor functions, which unify low-rank structures and implicit smooth regularization within a unified framework.

Specifically, LRTFR~\cite{luo2023low} substitutes the discrete factor matrices in Tucker decomposition with MLP-based continuous factor functions, achieving a theoretical unification of low-rank constraints and Lipschitz smoothness. DRO-TFF~\cite{li2025deep} adopts three separate INR parameterizations for each dimensional factor under a rank-1 spatial decomposition and deep modal transformation framework, enabling space-spectral dual smoothness encoding with extremely few parameters. CRNL~\cite{luo2024revisiting} leverages continuous functions instead of discrete block metrics to capture non-local self-similarity, and further exploits coupled low-rank function decomposition to model both cross-group shared patterns and intra-group individual variations. Furthermore, S-INR~\cite{li2024superpixel} incorporates superpixel semantic units with a joint-attention MLP and shared dictionaries. FA-INR~\cite{zhou2025frequency} performs divide-and-conquer modeling of frequency components in a learnable frequency space, thereby enhancing high-frequency representation capacity. Physics-guided priors have also shown effectiveness in inverse reconstruction, as demonstrated by MicroFM~\cite{zhan2026microfm}.

Although existing methods have achieved encouraging progress, they still fall short of addressing the optimization instability and insufficient spectral calibration based on natural-signal statistics in deep periodic INR from a fundamental architectural perspective, which is the key motivation of this work.

\section{Method}
\subsection{Our Motivations}
\label{sec:motivation}

Most INR methods based on periodic activation functions (e.g., sine~\cite{sitzmann2020implicit,saragadam2023wire}, Gabor wavelet~\cite{liu2024finer}) normalize input coordinates $\mathbf{v} \in [-1,1]^N$ (with $N$ being the signal dimensionality) and map them to signal values layer by layer, resulting in the following network output:
\begin{small}
\begin{equation}
\Phi_\theta(\mathbf{v}) = \phi_L \circ \cdots \circ \phi_1(\mathbf{v}), \quad 
\phi_l(\mathbf{x}) \triangleq \rho^{(l)}\!\left(\mathbf{W}^{(l)}\mathbf{x} + \mathbf{b}^{(l)}\right)
\end{equation}   
\end{small}

where $\mathbf{x}$ is the input to the layer, $\mathbf{W}^{(l)},\mathbf{b}^{(l)}$ are the weights and biases, and $\rho^{(l)}$ is the activation function. From this equation, the product of Jacobian matrices in backpropagation, $\mathbf{J}_\Phi = \prod_{l=1}^{L} \mathbf{J}_{\phi_l}$~\cite{glorot2010understanding}, leads to gradient vanishing or exploding as depth increases~\cite{lecun2015deep}, limiting network scalability.

From a harmonic perspective, the ideal signal representation should be the explicit superposition of frequency components, i.e., the generalized Fourier series:
\begin{small}
\begin{equation}
\Phi_\theta(\mathbf{v}) = \sum_k \Big[ \mathbf{A}_k \cos\big(\boldsymbol{\omega}_k^{\top} \mathbf{v}\big) + \mathbf{B}_k \sin\big(\boldsymbol{\omega}_k^{\top} \mathbf{v}\big) \Big]
\label{eq:fourier}
\end{equation}
\end{small}

where $\mathbf{A}_k,\mathbf{B}_k$ are the cosine and sine amplitudes of the $k$-th harmonic, and $\boldsymbol{\omega}_k$ are the angular frequencies. Notably, using the trigonometric identity $\sin(x+y)=\sin x\cos y+\cos x\sin y$, a sine unit with bias $\sin(\boldsymbol{\omega}^\top\mathbf{v}+b)$ can be decomposed as a linear combination of sine and cosine bases~\cite{Benbarka_2022_WACV}, making both representations fundamentally equivalent.

Comparing the composition structure $\phi_L \circ \cdots \circ \phi_1$ with Eq.~\eqref{eq:fourier}, we observe two major deficiencies in general INR architectures:
\begin{enumerate}[label=(\roman*)]
    \item \textbf{Lack of a superposition paradigm.} The nested function composition fails to establish a one-to-one correspondence between learnable coefficients and basis functions under the $\sum_k$ summation operator, making it difficult to achieve term-wise pairing with $\{(\mathbf{A}_k,\mathbf{B}_k)\}$. This impedes harmonic coexistence and tends to introduce conflicting behaviors that aggravate overfitting.
    \item \textbf{Frequency rigidity.} Common INRs adopt a globally fixed base frequency scaling (e.g., scaling all neuron activations by a constant $\omega_0$), resulting in a frequency support set entirely determined by initialization~\cite{yuce2022structured}, which is inconsistent with the $1/f$ power-law statistics of natural signals~\cite{field1987relations}. Furthermore, since the magnitude of parameter gradients is proportional to their corresponding frequencies, the update step of the gradient in the high frequency part is much larger than that of the low frequency, which is easy to produce shock and lead to unstable training.  
    
\end{enumerate}

\begin{figure*}[t!]
  \centering
  \includegraphics[width=\linewidth]{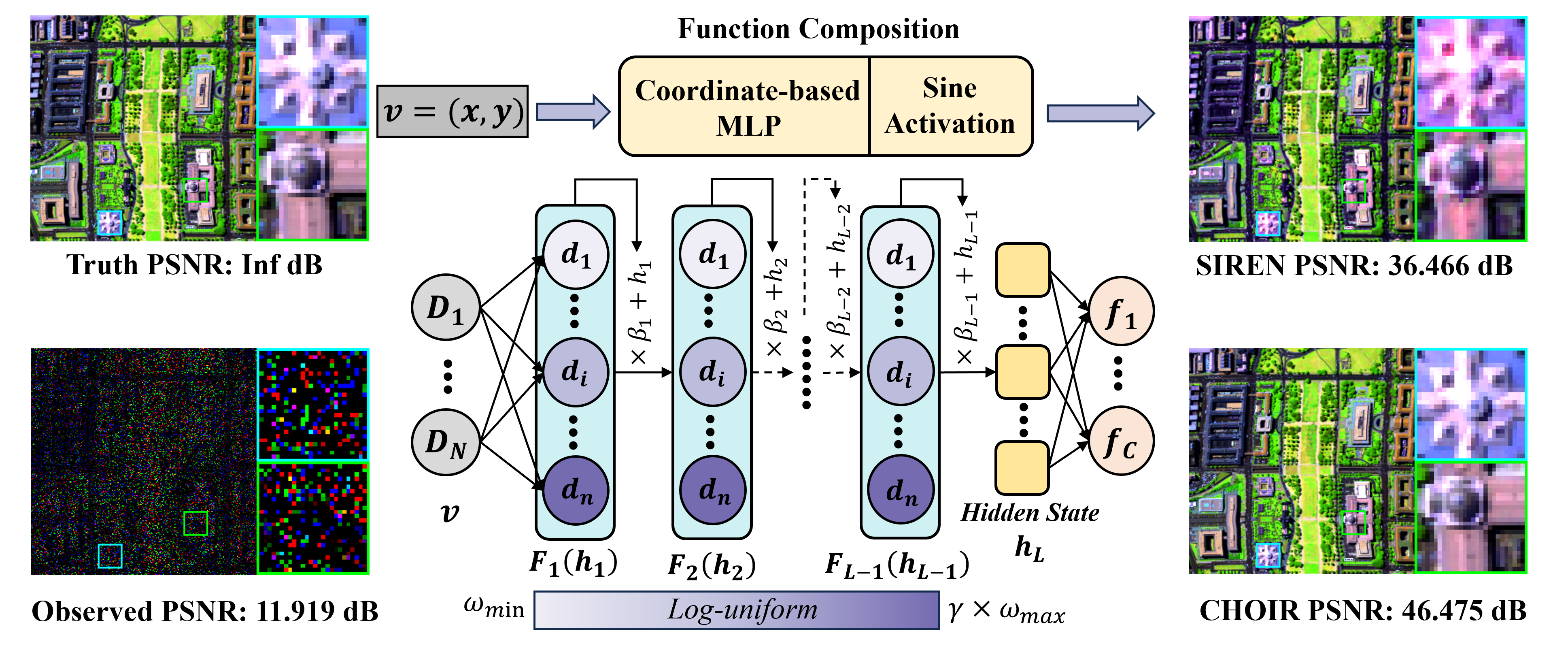}
  \caption{Overview of the architecture of our proposed CHOIR vs. most periodic INR methods (e.g., SIREN~\cite{sitzmann2020implicit}). CHOIR establishes a hybrid composition-superposition paradigm and leverages the power-law distribution of natural images for spectrum calibration. In the figure, different colors indicate different angular frequencies of neurons.}
  \label{fig:pipeline}
\end{figure*}

Based on these issues, although many INRs claim to integrate deep networks, experiments typically show optimal performance only in very shallow networks (3–5 layers), with rapid degradation as depth increases. Fourier harmonic analysis theory offers unique advantages in solving these two problems, motivating the work presented in this paper.
The overview of proposed CHOIR is shown in Figure~\ref{fig:pipeline}.

\subsection{Coordinated Harmonic Superposition (CHS)}
\label{sec:chs}

To address these issues, we propose the Coordinated Harmonic Superposition (CHS) mechanism, which restructures function composition into term-wise harmonic superposition. The network first maps the input coordinate $\mathbf{v}$ to an initial hidden state $\mathbf{h}_0(\mathbf{v}) = \mathbf{W}^{(0)}\mathbf{v} + \mathbf{b}^{(0)}$. Then, inspired by residual scaling~\cite{bachlechner2021rezero}, the output of each harmonic module $\mathcal{F}_l$ (detailed in Sec.~\ref{sec:psc}) is scaled by a learnable scalar $\beta_l$ (initialized to zero) and added to the input feature $\mathbf{h}_l$:
\begin{equation}
\mathbf{h}_{l+1} = \mathbf{h}_l + \beta_l \mathcal{F}_l(\mathbf{h}_l), \quad l = 0, 1, \dots, L-1
\label{eq:chs}
\end{equation}
Thus, at the beginning of training, the network degenerates into a simple linear mapping ($\mathbf{h}_{L} = \mathbf{h}_0$). This initialization ensures the overall Jacobian matrix $\mathbf{J}_\Phi$ is highly stable, providing a good starting point for optimization. The stability allows the network to first learn basic low-frequency components before handling higher‑frequency details.

This structure enables the network output to be expressed as an explicit additive superposition, corresponding to the generalized Fourier series:
\begin{small}
\begin{equation}
\Phi_\theta(\mathbf{v}) = \mathbf{W}_{\text{out}} \mathbf{h}_L(\mathbf{v}) = \mathbf{W}_{\text{out}} \left( \mathbf{h}_0(\mathbf{v}) + \sum_{l=0}^{L-1} \beta_l \mathcal{F}_l(\mathbf{h}_l(\mathbf{v})) \right)
\label{eq:output}
\end{equation}
\end{small}
where $\{\beta_l \mathbf{W}_{\text{out}} \mathcal{F}_l(\cdot)\}$ denotes a set of adaptive harmonic terms modulated by $\beta_l$ and $\mathbf{W}_{\text{out}}\,\mathbf{h}_0(\mathbf{v})$ corresponds to the zero‑frequency component. As noted in Sec.~\ref{sec:motivation}, a sine unit with bias can be decomposed into orthogonal sine and cosine bases, supporting harmonic superposition.

As optimization proceeds, the contribution of each harmonic term is dynamically introduced through its weight $\beta_l$. The gradient of $\beta_l$ is determined by the inner product between $\mathcal{F}_l(\mathbf{h}_l)$ and the downstream gradient $\frac{\partial \mathcal{L}}{\partial \mathbf{h}_{l+1}}$:
\begin{small}
\begin{equation}
\frac{\partial \mathcal{L}}{\partial \beta_l} = \left\langle \mathcal{F}_l(\mathbf{h}_l), \frac{\partial \mathcal{L}}{\partial \mathbf{h}_{l+1}} \right\rangle
\label{eq:beta_grad}
\end{equation}
\end{small}
This follows from the chain rule $\frac{\partial \mathcal{L}}{\partial \beta_l} = \frac{\partial \mathcal{L}}{\partial \mathbf{h}_{l+1}}^\top \frac{\partial \mathbf{h}_{l+1}}{\partial \beta_l}$ with $\frac{\partial \mathbf{h}_{l+1}}{\partial \beta_l} = \mathcal{F}_l(\mathbf{h}_l)$. Eq.~\eqref{eq:beta_grad} reveals an implicit curriculum learning mechanism~\cite{bengio2009curriculum} in CHS: the weight $\beta_l$ of a harmonic term increases only if its output direction $\mathcal{F}_l$ aligns with the negative gradient direction~\cite{kumar2010self}. This guides the network to begin with the simplest model and progressively activate the harmonic components most effective for fitting the residual, decomposing the complex optimization into a sequence of simpler subproblems.

\subsection{Perceptual Spectrum Calibration (PSC)}
\label{sec:psc}

To resolve frequency rigidity and specify the internal structure of $\mathcal{F}_l$, we further propose the Perceptual Spectrum Calibration (PSC). This mechanism embeds physical priors into the network via frequency sampling and amplitude modulation, achieving stable harmonic allocation that matches natural signal statistics.

\textbf{Log‑uniform Frequency Distribution.} We assign an angular frequency $\omega_{l,i}$ to each neuron in the $l$-th layer. The frequency range is set as $[\omega_{\min}, \omega_{\max}]$, with the lower bound $\omega_{\min} = \pi$, corresponding to the fundamental frequency on the normalized coordinate domain $[-1,1]^N$. According to the Nyquist sampling theorem~\cite{ricci2024nyquist}, the theoretical upper bound should be the Nyquist frequency \(\omega_{\text{Nyq}} = \pi \cdot \min(D_1,\dots,D_N)/2\). However, when taking it as the upper bound, the spatial period corresponding to this frequency contains only two sampling points, with a phase difference of $\pi$ between them. This causes the continuous sinusoid to degenerate into an alternating sequence of signs~\cite{smith2007mathematics}, leading to relatively unstable gradient optimization. Moreover, near this upper bound, the energy in natural signals is extremely low (following a $1/f$ spectrum~\cite{field1987relations}) and is sensitive to noise and sampling errors. Adopting it as the upper bound would waste capacity and increase the potential for error. Therefore we introduce a scaling factor $\gamma$ (determined by ablation in Sec.~\ref{sec:ablation}) and set

\begin{equation}
\omega_{\max} = \gamma \cdot \omega_{\text{Nyq}} = \gamma \cdot (\pi \cdot \min(D_1,\dots,D_N)/2).
\label{eq:Nyquist}
\end{equation}

Within the interval $[\omega_{\min}, \omega_{\max}]$, frequencies for $d$ neurons are assigned according to a geometric progression (i.e., equally spaced on a logarithmic axis):

\begin{equation}
\omega_{l,i} = \omega_{\min} \cdot \left(\frac{\omega_{\max}}{\omega_{\min}}\right)^{\frac{i-1}{d-1}}, \quad i = 1,\dots,d.
\label{eq:loguniform}
\end{equation}

Since adjacent neurons satisfy $\Delta\omega \propto \omega$, the sampling density becomes $\rho(\omega) \propto 1/\omega$, which matches the power‑law spectrum $P(\omega) \propto 1/\omega$ of natural signals~\cite{field1987relations}. Consequently, model capacity allocated to each frequency band is proportional to its energy, automatically assigning more parameters to energy‑dense low frequencies while maintaining sufficient high‑frequency coverage.

\textbf{Adaptive Amplitude Modulation.} To counteract the optimization instability caused by high‑frequency gradients dominating in standard periodic networks, we design an adaptive amplitude modulation method. Leveraging the power‑law spectrum decay $P(\omega)\propto \omega^{-\alpha}$, we constrain amplitudes to be proportional to the square root of the power spectrum density, i.e., $A \propto P^{1/2} \propto \omega^{-\alpha/2}$. Combined with log‑uniform sampling, the output of the $i$-th neuron in $\mathcal{F}_l$ is defined as:
\begin{small}
\begin{equation}
\mathcal{F}_l(\mathbf{h}_l)_i = \left(\frac{\tilde{\omega}_l}{\omega_{l,i}}\right)^{\alpha/2} \sin\big(\omega_{l,i}(\mathbf{w}_{l,i}^\top \mathbf{h}_l + b_{l,i})\big),
\label{eq:psc_output}
\end{equation}
\end{small}

where $\alpha$ is a globally learnable spectral decay rate (initialized to $2.0$), $\tilde{\omega}_l = \big(\prod_{j=1}^{d} \omega_{l,j}\big)^{1/d}$ is the geometric mean of frequencies in layer $l$, computed once at initialization and kept fixed. The amplitude factor $(\tilde{\omega}_l/\omega_{l,i})^{\alpha/2}$ compresses high‑frequency neuron outputs following the power law, suppressing their tendency to dominate gradients. Specifically, when $\alpha=2$, the gradient norm with respect to $\mathbf{w}_{l,i}$ becomes proportional to $\tilde{\omega}_l \cdot |\cos(\cdot)| \cdot \|\mathbf{h}_l\|_2$, which has the same expectation for all frequencies, yielding a naturally balanced training initialization. During training, $\alpha$ is adaptively updated to fit the actual spectral distribution of the target signal. In summary, PSC integrates physical priors through coordinated frequency and amplitude modulation, not only constraining the solution space but also improving the geometry of the optimization landscape, leading to stable and efficient training.

\begin{algorithm}[t!]
\caption{CHOIR Algorithm for Multi‑dimensional Data}
\label{alg:choir}
\begin{algorithmic}[1]
\REQUIRE Degraded observations $\mathcal{Y}$, mask $\Omega$, depth $L$, width $d$
\ENSURE Network parameters $\theta$
\STATE Pre-compute $\{\omega_{l,i}, \tilde{\omega}_l\}$ for all $l$ via log-uniform spacing (Eq.~\eqref{eq:Nyquist} Eq.~\eqref{eq:loguniform})
\STATE Initialize $\{\mathbf{W}^{(l)}, \mathbf{b}^{(l)}\}$ with SIREN scheme; $\beta_l \leftarrow 0$; $\alpha \leftarrow 2.0$
\FOR{epoch $= 1$ to MaxEpochs}
    \STATE $\mathbf{h}_0 \leftarrow \mathbf{W}^{(0)} \mathbf{v} + \mathbf{b}^{(0)}$
    \FOR{$l = 0$ to $L-1$}
        \STATE Compute $\mathcal{F}_l(\mathbf{h}_l)$ via PSC amplitude modulation (Eq.~\eqref{eq:psc_output})
        \STATE Update $\mathbf{h}_{l+1}$ via CHS additive gating (Eq.~\eqref{eq:chs})
    \ENDFOR
    \STATE $\Phi_\theta(\mathbf{v}) \leftarrow \mathbf{W}_{\text{out}} \mathbf{h}_L + \mathbf{b}_{\text{out}}$
    \STATE Update $\theta$ via Adam on $\mathcal{L}(\mathcal{P}_\Omega(\Phi_\theta), \mathcal{P}_\Omega(\mathcal{Y}))$
\ENDFOR
\RETURN $\theta$
\end{algorithmic}
\end{algorithm}

\subsection{Problem Formulation}
\label{sec:formulation}

To solve the inverse problem of multi‑dimensional data recovery, we employ an implicit neural representation (INR) as a self‑supervised prior. Concretely, we parameterize the target tensor $\mathcal{T}$ as a continuous mapping $\Phi_\theta: \mathbb{R}^N \to \mathbb{R}^C$ implemented by an MLP, where $N$ and $C$ are the coordinate dimensionality and number of channels, respectively. For any coordinate $\mathbf{v}\in[-1,1]^N$, the network outputs $\Phi_\theta(\mathbf{v})\in\mathbb{R}^C$. Querying the full coordinate grid produces the reconstructed tensor $\mathcal{T}(\theta)$. The recovery problem is then formulated as an optimization over the network parameters $\theta$:
\begin{equation}
\min_{\theta}\; \mathcal{L}(\theta) = \ell\!\left(\mathcal{P}_\Omega\big(\Phi_\theta\big),\; \mathcal{P}_\Omega(\mathcal{Y})\right),
\label{eq:objective}
\end{equation}
where $\mathcal{Y}$ is the degraded observation, $\mathcal{P}_\Omega$ is the projection operator that retains only entries in the observable index set $\Omega$, and $\ell(\cdot,\cdot)$ is a task‑adaptive per‑sample loss. The complete training procedure is summarized in Algorithm~\ref{alg:choir}. More analysis and results on convergence are provided in the supplementary material.




\section{Experiment}

\subsection{Experimental Setup}
We employ Peak Signal-to-Noise Ratio (PSNR, in dB), Structural Similarity Index (SSIM)~\cite{wang2003multiscale}, and Learned Perceptual Image Patch Similarity (LPIPS)~\cite{zhang2018unreasonable} to assess recovery quality. Additionally, the number of parameters (Params, in M) and runtime (Time, in seconds) are used to comprehensively evaluate all methods. All reported results are the average of five independent runs. In implementation, we use the Adam optimizer~\cite{kingma2014adam} with a learning rate of $1\times10^{-4}$ and weight decay. For all compared methods, we either use their open-source implementations or strictly reproduce them according to the published papers, carefully tuning hyperparameters to ensure optimal performance. All experiments are conducted on two NVIDIA RTX A6000 GPUs. In all tables of comparisons, the \textbf{best} and \underline{second-best} values are highlighted.

\subsection{Evaluation of Representation Capability}
\label{sec:representation}
To assess the intrinsic representation capability of INRs, we conduct two types of experiments: (1) \textbf{2D signal fitting}, which is commonly used as a benchmark for INRs due to its rich composition of both low and high-frequency components. We employ the House image ($512\times768\times3$) from the Kodak dataset~\cite{kodak1993} as the test data. (2) \textbf{5D Novel view synthesis}, which examines the capability of modeling neural radiance fields from 5D coordinates to RGB. We utilize the classic NeRF Blender synthetic dataset~\cite{mildenhall2021nerf}, which comprises 8 synthetic scenes with complex geometry and fine textures. From the training set, 25 images are randomly sampled (downsampled to $200\times200$) for training, and the synthesis quality on unseen viewpoints is evaluated. Comparison methods include: MLP with Positional Encoding (PEMLP)~\cite{tancik2020fourier}, SIREN~\cite{sitzmann2020implicit} based on sinusoidal activation functions, Gauss~\cite{ramasinghe2022beyond} which employs Gaussian activation functions to overcome purely periodic limitations, FINER~\cite{liu2024finer} which flexibly regulates spectral bias through variable-period activations, SL2A-INR~\cite{rezaeian2025sl2a} which introduces learnable single-layer activation functions, and IGA-INR~\cite{shi2025inductive} which mitigates spectral bias via induced gradient adjustment.

\begin{table}[b!]
\centering
\scriptsize
\setlength{\tabcolsep}{3pt}  
\renewcommand{\arraystretch}{0.90}  
\caption{Quantitative results on novel view synthesis in various methods.}
\label{tab:novel_view_synthesis}
\resizebox{\linewidth}{!}{
\begin{tabular}{c c *{8}{c}}
\toprule
& \textbf{Methods} & Chair & Drums & Ficus & Hotdog 
& Lego & Materials & Mic & Ship \\
\midrule

\multirow{7}{*}{\rotatebox[origin=c]{90}{PSNR\textuparrow}} 
 & PEMLP~\cite{mildenhall2021nerf}
 & 32.576 & 25.418 & 24.665 & 31.421 
 & \underline{25.340} & 25.319 & 28.281 & 25.277\\
 & SIREN~\cite{sitzmann2020implicit}
 & 32.843 & 24.719 & 24.949 & 35.187 
 & 23.793 & 27.114 & 32.394 & 22.719\\
 & Gauss~\cite{ramasinghe2022beyond}
 & 34.115 & 24.237 & 25.458 & \underline{36.555} 
 & 20.034 & 28.877 & 28.354 & 25.235\\
 & FINER~\cite{liu2024finer}
 & 32.249 & 24.553 & 25.472 & 33.901 
 & 23.700 & 26.516 & 31.633 & 25.625 \\
 & SL2A-INR~\cite{rezaeian2025sl2a}
 & 34.273 & 25.257 & \underline{25.539} & 36.088
 & 24.644 & \underline{29.698} & \underline{33.357} & 25.094\\
 & IGA-INR~\cite{shi2025inductive}
 & \underline{34.676} & \underline{25.997} & 24.891 & 36.066 
 & 23.550 & 28.131 & 31.584 & \underline{27.368}\\
 & CHOIR(Ours) 
 & \textbf{35.317} & \textbf{26.430} & \textbf{27.650} &\textbf{37.454} & \textbf{28.299} & \textbf{29.793} 
 & \textbf{35.613} & \textbf{27.555} \\
 \midrule
 
\multirow{7}{*}{\rotatebox[origin=c]{90}{SSIM\textuparrow}}
 & PEMLP~\cite{mildenhall2021nerf}
 & 0.972 & 0.921 & 0.904 & 0.959 
 & \underline{0.904} & 0.918 & 0.963 & 0.893\\
 & SIREN~\cite{sitzmann2020implicit}
 & 0.973 & 0.915 & 0.908 & 0.977 
 & 0.862 & 0.933 & 0.980 & 0.894\\
 & Gauss~\cite{ramasinghe2022beyond}
 & 0.979 & 0.893 & 0.909 & \underline{0.981} 
 & 0.768 & 0.944 & 0.964 & 0.895\\
 & FINER~\cite{liu2024finer}
 & 0.967 & 0.901 & 0.911 & 0.971 
 & 0.854 & 0.920 & 0.977 & 0.886\\
 & SL2A-INR~\cite{rezaeian2025sl2a}
 & 0.981 & 0.916 & \underline{0.922} & 0.979
 & 0.886 & \underline{0.951} & \underline{0.981} & 0.896\\
 & IGA-INR~\cite{shi2025inductive}
 & \underline{0.984} & \underline{0.930} & 0.903 & 0.980 
 & 0.884 & 0.937 & 0.973 & \underline{0.917}\\
 & CHOIR(Ours) 
 & \textbf{0.986} & \textbf{0.939} & \textbf{0.952} & \textbf{0.985} 
 & \textbf{0.933} & \textbf{0.955} & \textbf{0.987} & \textbf{0.926}\\
 \midrule

\multirow{7}{*}{\rotatebox[origin=c]{90}{LPIPS\textdownarrow}}
 & PEMLP~\cite{mildenhall2021nerf}
 & 0.036 & 0.097 & 0.161 & 0.048 
 & \textbf{0.077} & 0.086 & 0.050 & 0.093\\
 & SIREN~\cite{sitzmann2020implicit}
 & 0.034 & 0.103 & 0.153 & \underline{0.021} 
 & 0.130 & 0.053 & 0.020 & 0.134\\
 & Gauss~\cite{ramasinghe2022beyond}
 & 0.031 & 0.174 & 0.209 & \underline{0.021} 
 & 0.258 & 0.045 & 0.045 & 0.129\\
 & FINER~\cite{liu2024finer}
 & 0.040 & 0.112 & 0.175 & 0.030 
 & 0.150 & 0.055 & 0.019 & 0.130\\
 & SL2A-INR~\cite{rezaeian2025sl2a}
 & \underline{0.025} & 0.120 & \underline{0.110} & 0.024 
 & 0.110 & \underline{0.034} & \underline{0.015} & 0.084\\
 & IGA-INR~\cite{shi2025inductive}
 & 0.078 & \underline{0.090} & 0.144 & 0.042 
 & 0.094 & 0.036 & 0.064 & \textbf{0.069}\\
 & CHOIR(Ours) 
 & \textbf{0.017} & \textbf{0.088} & \textbf{0.078}
 & \textbf{0.015} & \underline{0.088} & \textbf{0.030} 
 & \textbf{0.013} & \underline{0.080}\\
 \bottomrule
\end{tabular}}
\end{table}

\begin{figure*}[t!]
  \centering
  \includegraphics[width=\linewidth]{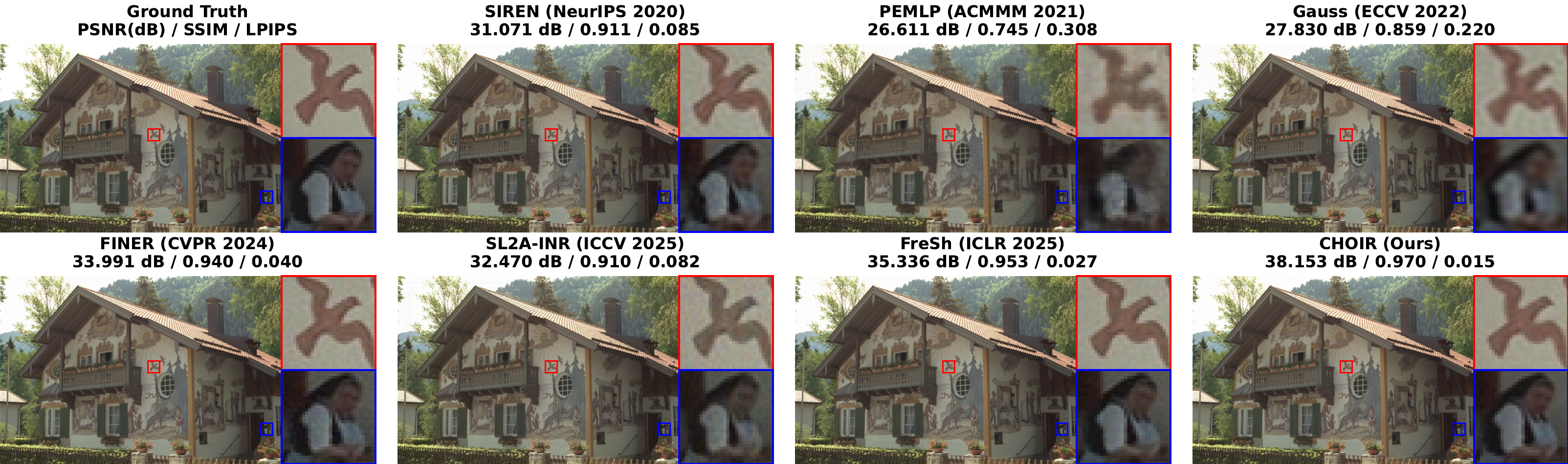}
  \caption{Results of signal fitting by different methods on RGB House dataset.}
  \label{fig:signal_fitting}
\end{figure*}

\begin{figure*}[t!]
  \centering
  \includegraphics[width=\linewidth]{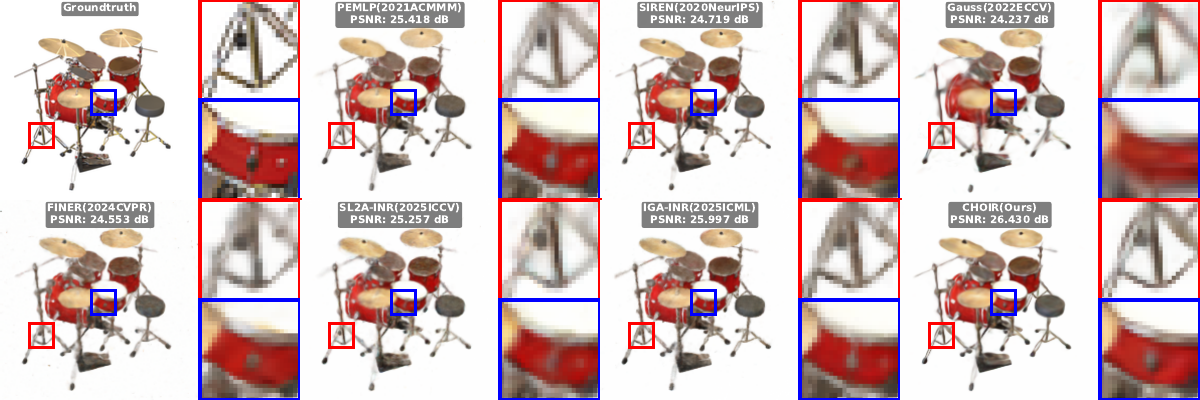}
  \caption{Results of novel view synthesis by different methods on Drums dataset.}
  \label{fig:nerf}
\end{figure*}

On the signal fitting task, as shown in Figure~\ref{fig:signal_fitting}, CHOIR consistently outperforms all comparison methods across all evaluation metrics (PSNR, SSIM, LPIPS) on the House dataset. Moreover, the visual results reveal that our method fits higher-frequency signals, such as the clearer contour of the woman under the eaves. For novel view synthesis, Table~\ref{tab:novel_view_synthesis} systematically demonstrates that CHOIR also achieves superior performance on the NeRF Blender dataset. From the zoomed-in details in Figure~\ref{fig:nerf}, it can be observed that CHOIR recovers the drums with more lustrous stands and sharper body. These two experiments validate the excellent fitting capability and generality of our proposed method.

\begin{table}[t!]
  \centering
  \scriptsize
  \setlength{\tabcolsep}{3pt}  
  \renewcommand{\arraystretch}{0.90}  
  \caption{Quantitative results on data missing completion in various methods.}
  \label{tab:missing_completion}
  \resizebox{\linewidth}{!}{
  \begin{tabular}{c c *{12}{c}}
    \toprule 
    \multirow{2}{*}{\textbf{Datasets}} &
    \multirow{2}{*}{\textbf{Methods}} & 
    \multicolumn{3}{c}{\textbf{Random OR=0.10}} &
    \multicolumn{3}{c}{\textbf{Random OR=0.30}} &
    \multicolumn{3}{c}{\textbf{Tube OR=0.10}} &
    \multicolumn{3}{c}{\textbf{Tube OR=0.30}} \\
    \cmidrule(lr){3-5} \cmidrule(lr){6-8} \cmidrule(lr){9-11} \cmidrule(lr){12-14}
    & & PSNR↑ & SSIM↑ & LPIPS↓ & PSNR↑ & SSIM↑ & LPIPS↓ & PSNR↑ & SSIM↑ & LPIPS↓ & PSNR↑ & SSIM↑ & LPIPS↓\\
    \midrule
      \multirow{7}{*}{\begin{tabular}[c]{@{}c@{}}HSI\\WDC\end{tabular}}
      & SIREN\cite{sitzmann2020implicit} 
      & 36.466 & 0.901 & 0.032 & 37.149 & 0.905 & 0.029 
      & 24.670 & 0.661 & 0.260 & 27.832 & 0.783 & 0.134\\
      & LRTFR\cite{luo2023low} 
      & 32.530 & 0.917 & 0.121 & 34.201 & 0.942 & 0.069
      & 23.643 & 0.648 & 0.367 & 27.983 & 0.834 & 0.168\\
      & CRNL\cite{luo2024revisiting} 
      & \underline{44.212} & \underline{0.994} & \underline{0.009}
      & \underline{50.694} & \underline{0.998} & \underline{0.003}
      & \underline{27.158} & \underline{0.803} & \underline{0.246} 
      & \underline{31.207} & \underline{0.919} & 0.089\\
      & FINER\cite{liu2024finer} 
      & 43.016 & 0.980 & 0.010 & 46.090 & 0.988 & 0.007 
      & 24.390 & 0.696 & 0.374 & 27.392 & 0.818 & 0.214\\
      & DRO-TFF\cite{li2025deep} 
      & 40.254 & 0.981 & 0.015 & 42.084 & 0.985 & 0.009 
      & 26.152 & 0.743 & 0.387 & 30.861 & 0.904 & \underline{0.085}\\
      & FreSh\cite{kania2025fresh} 
      & 38.762 & 0.916 & 0.020 & 40.145 & 0.922 & 0.017
      & 23.310 & 0.607 & 0.428 & 25.883 & 0.713 & 0.265\\
      & CHOIR 
      & \textbf{46.475} & \textbf{0.996} & \textbf{0.004}
      & \textbf{51.359} & \textbf{0.999} & \textbf{0.002}
      & \textbf{28.142} & \textbf{0.832} & \textbf{0.196}
      & \textbf{31.414} & \textbf{0.928} & \textbf{0.080}\\
    \midrule
      \multirow{7}{*}{\begin{tabular}[c]{@{}c@{}}HSI\\Urban\end{tabular}}
      & SIREN\cite{sitzmann2020implicit}
      & 39.482 & 0.984 & 0.012 & 40.835 & \underline{0.988} & 0.010 
      & 21.756 & 0.562 & 0.272 & 25.138 & 0.782 & 0.109 \\
      & LRTFR\cite{luo2023low} 
      & 29.979 & 0.978 & 0.135 & 32.479 & 0.933 & 0.063
      & 19.478 & 0.315 & 0.499 & 24.617 & 0.700 & 0.218\\
      & CRNL\cite{luo2024revisiting} 
      & \underline{42.533} & \underline{0.993} & \underline{0.006} 
      & \underline{47.955} & \textbf{0.998} & \underline{0.003} 
      & \underline{23.423} & \underline{0.638} & \underline{0.255}
      & 27.343 & 0.840 & \textbf{0.091}\\
      & FINER\cite{liu2024finer} 
      & 38.495 & 0.979 & 0.013 & 39.504 & 0.982 & 0.011
      & 21.440 & 0.494 & 0.365 & 24.692 & 0.716 & 0.184\\
      & DRO-TFF\cite{li2025deep} 
      & 37.150 & 0.979 & 0.014 & 38.671 & 0.984 & 0.009
      & 23.023 & 0.601 & 0.293 
      & \underline{27.357} & \underline{0.843} & \underline{0.096}\\
      & FreSh\cite{kania2025fresh} 
      & 38.974 & 0.982 & 0.013 & 39.791 & 0.985 & 0.011
      & 20.242 & 0.440 & 0.416 & 23.800 & 0.692 & 0.254\\
      & CHOIR 
      & \textbf{44.412} & \textbf{0.995} & \textbf{0.005}
      & \textbf{49.317} & \textbf{0.998} & \textbf{0.001}
      & \textbf{23.493} & \textbf{0.643} & \textbf{0.247} 
      & \textbf{27.578} & \textbf{0.847} & \underline{0.096}\\
    \midrule
      \multirow{7}{*}{\begin{tabular}[c]{@{}c@{}}MSI\\Flowers\end{tabular}}
      & SIREN\cite{sitzmann2020implicit}  
      & 37.268 & 0.854 & 0.036 & 38.570 & 0.863 & 0.030 
      & \underline{28.159} & 0.764 & \textbf{0.119} & 32.992 & 0.836 & 0.054\\
      & LRTFR\cite{luo2023low} 
      & 33.735 & 0.892 & 0.160 & 35.660 & 0.933 & 0.117
      & 25.337 & 0.711 & 0.337 & 31.677 & 0.884 & 0.176\\
      & CRNL\cite{luo2024revisiting} 
      & 39.975 & \underline{0.991} & 0.031 & 
      \underline{48.604} & \textbf{0.998} & \underline{0.005} 
      & 27.333 & \underline{0.882} & 0.182 
      & 33.097 & \underline{0.964} & 0.053\\
      & FINER\cite{liu2024finer} 
      & 37.995 & 0.940 & 0.051 & 45.436 & 0.971 & 0.006 
      & 24.833 & 0.743 & 0.237 & 31.814 & 0.869 & 0.078\\
      & DRO-TFF\cite{li2025deep} 
      & \underline{40.912} & 0.986 & \underline{0.017}
      & 45.754 & \underline{0.996} & \underline{0.005}
      & 27.991 & 0.871 & 0.146 
      & \underline{33.277} & \underline{0.964} & \underline{0.051}\\
      & FreSh\cite{kania2025fresh} 
      & 36.660 & 0.855 & 0.040 & 39.569 & 0.869 & 0.021
      & 27.096 & 0.735 & 0.166 & 31.898 & 0.824 & 0.057\\
      & CHOIR 
      & \textbf{41.316} & \textbf{0.992} & \textbf{0.014}
      & \textbf{50.395} & \textbf{0.998} & \textbf{0.001} 
      & \textbf{28.726} & \textbf{0.894} & \underline{0.138}
      & \textbf{33.465} & \textbf{0.973} & \textbf{0.048}\\
    \midrule
      \multirow{7}{*}{\begin{tabular}[c]{@{}c@{}}MSI\\Toys\end{tabular}}
      & SIREN\cite{sitzmann2020implicit}  
      & 37.147 & 0.927 & 0.035 & 38.964 & 0.934 & 0.029 
      & \underline{25.547} & 0.827 
      & \textbf{0.182} & 30.595 & 0.906 & 0.058\\
      & LRTFR\cite{luo2023low} 
      & 32.135 & 0.907 & 0.144 & 34.115 & 0.943 & 0.082
      & 22.771 & 0.707 & 0.398 & 28.963 & 0.884 & 0.170\\
      & CRNL\cite{luo2024revisiting} 
      & 39.582 & \underline{0.992} & 0.018 
      & \underline{49.015} & \textbf{0.998} & \underline{0.003}
      & 24.908 & \underline{0.853} & 0.212 
      & \underline{30.883} & \underline{0.955} & \underline{0.054}\\
      & FINER\cite{liu2024finer} 
      & 37.430 & 0.964 & 0.051 & 46.818 & 0.989 & 0.005 
      & 22.863 & 0.775 & 0.316 & 28.435 & 0.918 & 0.122\\
      & DRO-TFF\cite{li2025deep} 
      & \underline{39.846} & \underline{0.992} & \underline{0.013}
      & 43.610 & \underline{0.996} & 0.004 
      & 25.103 & 0.851 & 0.217 & 30.728 & 0.951 & 0.061\\
      & FreSh\cite{kania2025fresh} 
      & 36.553 & 0.927 & 0.052 & 40.018 & 0.939 & 0.025
      & 24.443 & 0.785 & 0.267 & 29.580 & 0.893 & 0.089\\
      & CHOIR 
      & \textbf{41.748} & \textbf{0.993} & \textbf{0.010} 
      & \textbf{49.216} & \textbf{0.998} & \textbf{0.002}
      & \textbf{25.640} & \textbf{0.864} & \underline{0.211} 
      & \textbf{30.903} & \textbf{0.958} & \textbf{0.051}\\
    \midrule
      \multirow{7}{*}{\begin{tabular}[c]{@{}c@{}}Video\\Shop\end{tabular}}
      & SIREN\cite{sitzmann2020implicit}
      & 28.146 & 0.854 & 0.123 & \underline{31.551} & 0.933 & 0.061 
      & 21.087 & 0.577 & 0.485 & 25.704 & 0.861 & 0.168\\
      & LRTFR\cite{luo2023low} 
      & 25.567 & 0.864 & 0.182 & 27.667 & 0.902 & 0.118
      & 20.229 & 0.599 & 0.449 & 24.671 & 0.844 & 0.179\\
      & CRNL\cite{luo2024revisiting} 
      & 27.877 & 0.904 & 0.094 & 29.755 & 0.932 & 0.069 
      & 21.150 & \underline{0.774} & \underline{0.285} 
      & \underline{26.390} & \underline{0.907} & 0.109\\
      & FINER\cite{liu2024finer} 
      & 24.470 & 0.679 & 0.320 & 30.914 & 0.902 & 0.081 
      & 16.171 & 0.203 & 0.969 & 23.848 & 0.698 & 0.378\\
      & DRO-TFF\cite{li2025deep} 
      & \underline{28.729} & \underline{0.927} & \underline{0.060}
      & 29.376 & \underline{0.936} & \underline{0.056}
      & \underline{22.338} & 0.759 & 0.286
      & 26.362 & 0.900 & \underline{0.100}\\
      & FreSh\cite{kania2025fresh} 
      & 26.564 & 0.788 & 0.188 & 31.451 & 0.924 & 0.068 
      & 18.762 & 0.376 & 0.712 & 24.433 & 0.767 & 0.338\\
      & CHOIR 
      & \textbf{29.383} & \textbf{0.942} & \textbf{0.034}
      & \textbf{33.581} & \textbf{0.974} & \textbf{0.016} 
      & \textbf{23.267} & \textbf{0.781} & \textbf{0.263}
      & \textbf{26.656} & \textbf{0.918} & \textbf{0.099}\\
    \midrule
      \multirow{7}{*}{\begin{tabular}[c]{@{}c@{}}Video\\News \end{tabular}}
      & SIREN\cite{sitzmann2020implicit}  
      & 31.502 & 0.933 & 0.047 & 33.586 & 0.947 & 0.036 
      & 20.395 & 0.731 & 0.231 & 26.120 & 0.906 & 0.053\\
      & LRTFR\cite{luo2023low} 
      & 28.188 & 0.855 & 0.149 & 30.566 & 0.901 & 0.084
      & 18.684 & 0.459 & 0.482 & 24.820 & 0.757 & 0.250\\
      & CRNL\cite{luo2024revisiting} 
      & 31.618 & 0.953 & 0.036 
      & \underline{36.686} & \underline{0.983} & \underline{0.011}
      & 21.825 & 0.716 & 0.286 
      & 26.390 & \underline{0.918} & 0.099\\
      & FINER\cite{liu2024finer} 
      & 30.414 & 0.894 & 0.056 & 36.631 & 0.958 & 0.012 
      & 19.819 & 0.554 & 0.385 & 25.170 & 0.805 & 0.145\\
      & DRO-TFF\cite{li2025deep} 
      & \underline{32.505} & \underline{0.955} & \underline{0.025}
      & 35.557 & 0.969 & 0.014
      & \underline{22.148} & \underline{0.805} & \underline{0.227} 
      & \underline{27.402} & 0.917 & \underline{0.075}\\
      & FreSh\cite{kania2025fresh} 
      & 29.532 & 0.912 & 0.062 & 34.337 & 0.953 & 0.017 
      & 19.054 & 0.528 & 0.430 & 25.408 & 0.845 & 0.117\\
      & CHOIR 
      & \textbf{33.502} & \textbf{0.963} & \textbf{0.021} 
      & \textbf{38.739} & \textbf{0.986} & \textbf{0.006} 
      & \textbf{22.597} & \textbf{0.818} & \textbf{0.166}
      & \textbf{27.518} & \textbf{0.920} & \textbf{0.069}\\
    \bottomrule
  \end{tabular}}
\end{table}

\subsection{Evaluation on Data Recovery Task}

To assess the performance of CHOIR on multi-dimensional data recovery tasks, we conduct two categories of experiments: (1) \textbf{Data Missing Completion}, used to evaluate the capability of recovering complete data from partial observations. Experiments are performed under two observation rates ($OR \in \{10\%, 30\%\}$), considering both random missing and tube missing patterns. (2) \textbf{Mixed Degradation Restoration}, used to evaluate the recovery capability under coexisting complex noise and non-uniform missing data. Three progressive scenarios are designed: \textbf{Scene 1} adds Gaussian noise ($\sigma = 0.20$); \textbf{Scene 2} further adds salt-and-pepper noise with a sampling rate ($SR=10\%$); \textbf{Scene 3} additionally removes $3\%$ of entire rows and $3\%$ of entire columns from all channels to simulate sensor stripe failures. The datasets used in these two categories cover various modalities of multi-dimensional data: (1) \textbf{Hyperspectral Images (HSI)}: Pavia University (cropped to $200\times200\times80$)~\cite{liu2012tensor}, Washington DC Mall (cropped to $256\times256\times191$)~\cite{liu2012tensor}, and Urban (downsampled to $256\times256\times162$); (2) \textbf{Multispectral Images (MSI)}: Cloth, Toys, and Flowers from the CAVE dataset~\cite{yasuma2010generalized} (all downsampled to $256\times256\times31$); (3) \textbf{Videos}: Grayscale video Shop ($144\times192\times157$)~\cite{li2004statistical} and RGB video News ($288\times352\times3\times10$)~\cite{bengua2017efficient}; (4) \textbf{RGB Images}: Bird and Statue from the Kodak dataset~\cite{kodak1993} (downsampled to $256\times256\times3$). In addition to representative INR methods, we also include continuous representation methods that integrate structured priors with INRs as baselines. These include SIREN~\cite{sitzmann2020implicit}, FINER~\cite{liu2024finer}, FreSh~\cite{kania2025fresh} (which initializes frequency offsets to match the target signal spectrum), LRTFR~\cite{luo2023low} (which incorporates functional transformations into a low-rank factorization framework), CRNL~\cite{luo2024revisiting} (which exploits non-local self-similarity priors via continuous representations), and DRO-TFF~\cite{li2025deep} (based on depth-one tensor function factorization). Notably, in mixed degradation experiments, continuous representation methods such as LRTFR employ TV regularization following their original settings to strengthen denoising and smoothing. In contrast, CHOIR inherently encodes strong implicit smoothing priors and does not require task-specific explicit regularization such as TV regularization.

\begin{figure*}[t!]
  \centering
  \includegraphics[width=\linewidth]{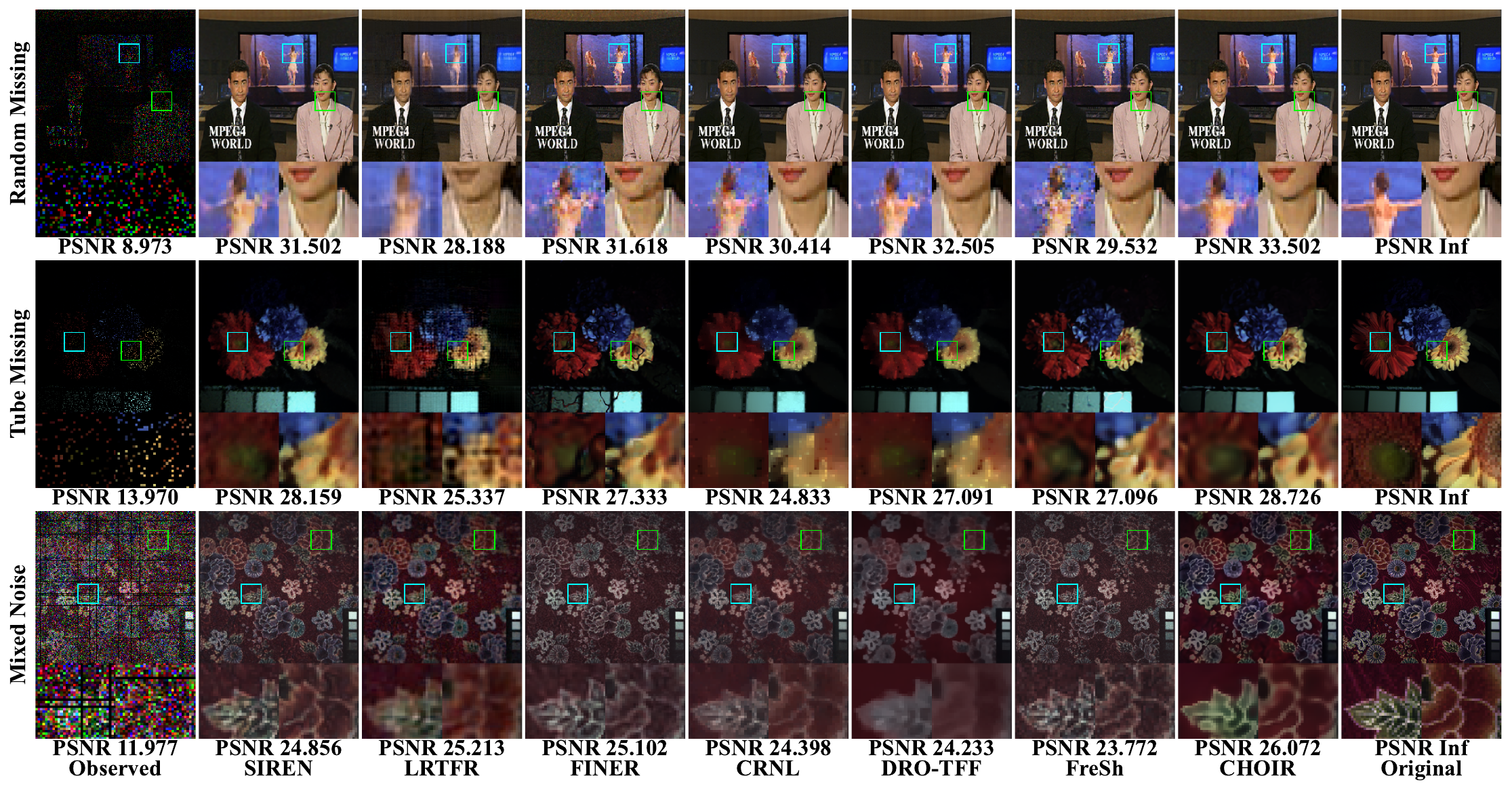}
  \caption{Results of different methods on data recovery tasks. Top: random missing and OR=0.1 on Video News dataset. Middle: tube missing and OR=0.1 on MSI Flowers dataset. Bottom: mixed degradation restoration (Scene 3) on MSI Cloth dataset.}
  \label{fig:recovery}
\end{figure*}

Tables~\ref{tab:missing_completion} and~\ref{tab:mixed_degradation} systematically demonstrate that CHOIR achieves the best performance on both categories of multi-dimensional data recovery experiments. This consistent advantage in scenarios with large-scale missing entries or dense noise with non-uniform structured missing entries strongly validates the detail preservation and recovery capabilities of CHOIR. The recovery of various multi-dimensional data (e.g., RGB, HSI, Video) also systematically reflects the generalization of CHOIR. As shown in Figure~\ref{fig:recovery}, the visual results indicate that our method exhibits robust performance against uniform missing, non-uniform missing, and complex noise. For instance, the lip edges in News are smoother, the petal details in Flowers are more realistic, and the pattern colors in Cloth are more lustrous. Due to space limitations, more experimental visualization results are detailed in the supplementary materials, including additional high-resolution experiments on the full CAVE dataset ($512\times512\times31$).

\begin{table}[t!]
  \centering
  \scriptsize
  \setlength{\tabcolsep}{2pt}  
  \renewcommand{\arraystretch}{0.90}  
  \caption{Quantitative results under mixed degradation scenario in various methods.}
  \label{tab:mixed_degradation}
  \resizebox{\linewidth}{!}{
  \begin{tabular}{c c *{9}{c}}
    \toprule 
    \multirow{2}{*}{\textbf{Datasets}} &
    \multirow{2}{*}{\textbf{Methods}} & 
    \multicolumn{3}{c}{\textbf{Scene 1}} &
    \multicolumn{3}{c}{\textbf{Scene 2}} &
    \multicolumn{3}{c}{\textbf{Scene 3}} \\
    \cmidrule(lr){3-5} \cmidrule(lr){6-8} \cmidrule(lr){9-11}
    & & PSNR↑ & SSIM↑ & LPIPS↓ & PSNR↑ & SSIM↑ & LPIPS↓ & PSNR↑ & SSIM↑ & LPIPS↓\\
    \midrule
      \multirow{8}{*}{\begin{tabular}[c]{@{}c@{}}HSI\\WDC\end{tabular}}
      & SIREN\cite{sitzmann2020implicit} 
      & 24.729 & 0.588 & 0.134 
      & 24.688 & 0.585 & 0.142 
      & 24.585 & 0.578 & 0.149\\
      & LRTFR\cite{luo2023low} 
      & \underline{28.615} & \underline{0.610} & 0.292 
      & \underline{28.468} & 0.607 & 0.302 
      & \underline{28.255} & 0.591 & 0.308\\
      & CRNL\cite{luo2024revisiting} 
      & 23.373 & 0.435 & 0.574 & 23.602 & 0.464 & 0.538 
      & 23.529 & 0.459 & 0.489\\
      & FINER\cite{liu2024finer} 
      & 24.766 & 0.585 & \underline{0.128} & 24.774 & 0.586 & \underline{0.130}
      & 24.669 & 0.575 & 0.146\\
      & DRO-TFF\cite{li2025deep} 
      & 24.188 & 0.518 & 0.340 & 25.532 & \underline{0.635} & 0.521 
      & 25.527 & \underline{0.634} & 0.499\\
      & FreSh\cite{kania2025fresh} 
      & 24.646 & 0.582 & 0.144 & 24.620 & 0.578 & \underline{0.146}
      & 24.423 & 0.567 & 0.152\\
      & CHOIR
      & \textbf{31.242} & \textbf{0.884} & \textbf{0.118} 
      & \textbf{31.047} & \textbf{0.883} & \textbf{0.126} 
      & \textbf{30.595} & \textbf{0.870} & \textbf{0.145}\\
    \midrule
      \multirow{8}{*}{\begin{tabular}[c]{@{}c@{}}HSI\\Pavia\end{tabular}}
      & SIREN\cite{sitzmann2020implicit}  
      & 27.641 & 0.870 & \underline{0.059} 
      & 27.507 & \underline{0.867} & \underline{0.062} 
      & 27.228 & 0.853 & \underline{0.068}\\
      & LRTFR\cite{luo2023low} 
      & \underline{28.212} & 0.824 & 0.182 
      & \underline{28.080} & 0.819 & 0.189 
      & \underline{27.847} & 0.812 & 0.194\\
      & CRNL\cite{luo2024revisiting} 
      & 27.103 & 0.820 & 0.093 & 27.028 & 0.816 & 0.095 
      & 26.790 & 0.810 & 0.098\\
      & FINER\cite{liu2024finer} 
      & 27.501 & \underline{0.871} & 0.082 & 27.392 & 0.865 & 0.083 
      & 27.107 & \underline{0.854} & 0.089\\
      & DRO-TFF\cite{li2025deep} 
      & 27.323 & 0.836 & 0.095 & 26.105 & 0.730 & 0.224 
      & 25.865 & 0.718 & 0.242\\
      & FreSh\cite{kania2025fresh} 
      & 27.056 & 0.855 & 0.076 & 26.756 & 0.839 & 0.075
      & 26.718 & 0.829 & 0.107\\
      & CHOIR
      & \textbf{30.401} & \textbf{0.904} & \textbf{0.048}
      & \textbf{29.959} & \textbf{0.895} & \textbf{0.057}
      & \textbf{29.691} & \textbf{0.882} & \textbf{0.047}\\
    \midrule
      \multirow{8}{*}{\begin{tabular}[c]{@{}c@{}}MSI\\Flowers\end{tabular}}
      & SIREN\cite{sitzmann2020implicit} 
      & 23.400 & 0.470 & 0.187 & 23.356 & 0.471 & \underline{0.195}
      & 23.315 & 0.465 & \underline{0.201}\\
      & LRTFR\cite{luo2023low} 
      & \underline{29.749} & \underline{0.601} & 0.227 & 29.676 & 0.600 & 0.237 
      & 29.528 & 0.596 & 0.237\\
      & CRNL\cite{luo2024revisiting} 
      & 23.822 & 0.465 & 0.297 & 23.766 & 0.464 & 0.205 
      & 23.624 & 0.455 & 0.370\\
      & FINER\cite{liu2024finer} 
      & 23.345 & 0.462 & 0.224 & 23.287 & 0.456 & 0.236
      & 23.204 & 0.452 & 0.245\\
      & DRO-TFF\cite{li2025deep} 
      & 23.861 & 0.469 & \underline{0.154} 
      & \underline{30.151} & \underline{0.701} & 0.214 
      & \underline{29.805} & \underline{0.696} & 0.221\\
      & FreSh\cite{kania2025fresh} 
      & 23.068 & 0.452 & 0.228 & 22.975 & 0.465 & 0.272 
      & 22.956 & 0.451 & 0.218\\
      & CHOIR
      & \textbf{33.094} & \textbf{0.856} & \textbf{0.122}
      & \textbf{32.919} & \textbf{0.852} & \textbf{0.129}
      & \textbf{32.605} & \textbf{0.847} & \textbf{0.130}\\
    \midrule
      \multirow{8}{*}{\begin{tabular}[c]{@{}c@{}}MSI\\Cloth\end{tabular}}
      & SIREN\cite{sitzmann2020implicit} 
      & 25.435 & 0.774 & \underline{0.164} 
      & 25.250 & \underline{0.768} & 0.174 
      & 24.856 & 0.753 & 0.182\\
      & LRTFR\cite{luo2023low} 
      & \underline{26.076} & 0.733 & 0.216 
      & \underline{25.953} & 0.727 & 0.221 
      & \underline{25.213} & 0.716 & 0.229\\
      & CRNL\cite{luo2024revisiting} 
      & 25.265 & \underline{0.783} & 0.201 & 25.301 & 0.765 & 0.194 
      & 25.102 & \underline{0.777} & 0.205\\
      & FINER\cite{liu2024finer} 
      & 24.826 & 0.753 & 0.268 & 24.504 & 0.760 & \underline{0.161}
      & 24.398 & 0.752 & \underline{0.173}\\
      & DRO-TFF\cite{li2025deep} 
      & 25.799 & 0.726 & 0.171 & 24.584 & 0.739 & 0.261 
      & 24.233 & 0.717 & 0.262\\
      & FreSh\cite{kania2025fresh} 
      & 24.422 & 0.756 & 0.212 & 24.287 & 0.749 & 0.220
      & 23.772 & 0.727 & 0.237\\
      & CHOIR
      & \textbf{26.612} & \textbf{0.798} & \textbf{0.138} 
      & \textbf{26.482} & \textbf{0.794} & \textbf{0.147} 
      & \textbf{26.072} & \textbf{0.779} & \textbf{0.152}\\
    \midrule
      \multirow{8}{*}{\begin{tabular}[c]{@{}c@{}}RGB\\Bird\end{tabular}}
      & SIREN\cite{sitzmann2020implicit}
      & \underline{24.639} & \underline{0.695} & \underline{0.324}
      & \underline{24.505} & \underline{0.679} & \underline{0.344} 
      & \underline{24.345} & \underline{0.667} & \underline{0.355}\\
      & LRTFR\cite{luo2023low} 
      & 17.217 & 0.357 & 0.808 & 17.089 & 0.336 & 0.822 
      & 16.986 & 0.323 & 0.824\\
      & CRNL\cite{luo2024revisiting} 
      & 20.914 & 0.338 & 0.710 & 20.009 & 0.304 & 0.752 
      & 19.482 & 0.286 & 0.784\\
      & FINER\cite{liu2024finer} 
      & 22.793 & 0.680 & 0.391 & 22.646 & 0.664 & 0.404 
      & 22.655 & 0.664 & 0.407\\
      & DRO-TFF\cite{li2025deep} 
      & 19.046 & 0.277 & 0.895 & 17.460 & 0.626 & 0.751 
      & 17.479 & 0.647 & 0.744\\
      & FreSh\cite{kania2025fresh} 
      & 23.015 & 0.674 & 0.400 & 22.864 & 0.653 & 0.422
      & 22.797 & 0.647 & 0.435\\
      & CHOIR 
      & \textbf{25.536} & \textbf{0.736} & \textbf{0.311} 
      & \textbf{25.380} & \textbf{0.724} & \textbf{0.325} 
      & \textbf{25.115} & \textbf{0.720} & \textbf{0.335}\\
    \midrule
      \multirow{8}{*}{\begin{tabular}[c]{@{}c@{}}RGB\\Statue\end{tabular}}
      & SIREN\cite{sitzmann2020implicit} 
      & \underline{24.464} & \underline{0.681} & 0.481
      & \underline{24.350} & \underline{0.673} & \underline{0.488}
      & \underline{24.220} & \underline{0.666} & \underline{0.492}\\
      & LRTFR\cite{luo2023low} 
      & 22.391 & 0.487 & 0.635 & 22.012 & 0.464 & 0.656 
      & 21.680 & 0.445 & 0.662\\
      & CRNL\cite{luo2024revisiting} 
      & 18.976 & 0.317 & 0.958 & 18.388 & 0.293 & 1.033
      & 18.171 & 0.285 & 1.037\\
      & FINER\cite{liu2024finer} 
      & 23.547 & 0.618 & 0.494 & 23.225 & 0.598 & 0.498 
      & 22.820 & 0.574 & 0.494\\
      & DRO-TFF\cite{li2025deep} 
      & 21.231 & 0.446 & 0.763 & 24.118 & 0.650 & 0.524 
      & 24.100 & 0.651 & 0.527\\
      & FreSh\cite{kania2025fresh} 
      & 22.990 & 0.584 & \underline{0.477} & 22.701 & 0.633 & 0.493 
      & 22.628 & 0.627 & 0.501\\
      & CHOIR  
      & \textbf{25.534} & \textbf{0.702} & \textbf{0.446} 
      & \textbf{25.456} & \textbf{0.700} & \textbf{0.470}
      & \textbf{25.175} & \textbf{0.679} & \textbf{0.481}\\
    \bottomrule
  \end{tabular}}
\end{table}

\subsection{Ablation Studies}
\label{sec:ablation}
\textbf{Effectiveness of Core Components of Our Method.} To validate the effectiveness of the core components of our method, we conduct ablation studies on the MSI Flowers dataset. As summarized in Table~\ref{tab:ablation}, introducing CHS alone yields a significant PSNR improvement over the sinusoidal activation baseline, demonstrating that its additive compositional structure effectively stabilizes deep optimization. While employing PSC alone, reducing the frequency scaling factor $\gamma$ from 1.0 to $1/8$ brings consistent gains, highlighting the superiority of embedding the $1/f$ power-law prior and frequency scaling factor $\gamma$. However, an excessively small $\gamma$ (e.g., $1/16$) impedes the model from capturing necessary high-frequency components, leading to performance degradation. Therefore, the experiments use $\gamma=1/8$. The full CHOIR model, which integrates both CHS and PSC, achieves the best performance across all evaluation metrics, verifying the effective synergy between these two core components.
\begin{table}[t!]
  \centering
  \scriptsize
  \setlength{\tabcolsep}{3pt}  
  \caption{Ablation studies of our core components on MSI Flowers dataset.}
  \label{tab:ablation}
  \begin{tabular}{c c *{6}{c} c c}
    \toprule
      \multirow{2}{*}{\textbf{Variants}} & \multirow{2}{*}{\textbf{Scale $\gamma$}} 
      & \multicolumn{3}{c}{\textbf{Random OR=0.10}} 
      & \multicolumn{3}{c}{\textbf{Scene 3}} 
      & \multirow{2}{*}{\textbf{Params}}
      & \multirow{2}{*}{\textbf{Time/iter}}
      \\
    \cmidrule(lr){3-5} \cmidrule(lr){6-8}
    & & PSNR↑ & SSIM↑ & LPIPS↓ & PSNR↑ & SSIM↑ & LPIPS↓\\
    \midrule
     Sine & - & 33.500 & 0.929 & 0.099 & 26.238 & 0.487 & 0.363
     & 0.202M & 0.046s\\
     Sine+CHS & - & 39.136 & \underline{0.982} & 0.018 
     & \underline{32.296} & \underline{0.801} & \underline{0.135} 
     & 0.202M & 0.054s\\
     Sine+PSC & 1.0 & 37.251 & 0.969 & 0.044 & 28.025 & 0.599 & 0.273
     & 0.204M & 0.050s\\
     Sine+PSC & 1/2 & 37.507 & 0.968 & 0.036 & 29.094 & 0.627 & 0.262
     & 0.204M & 0.050s\\
     Sine+PSC & 1/4 & 39.846 & 0.980 & 0.019 & 30.308 & 0.691 & 0.159
     & 0.204M & 0.050s\\
     Sine+PSC & \textbf{1/8} & \underline{40.335} & \underline{0.982} & \underline{0.016} & 31.143 & 0.704 & 0.147 
     & 0.204M & 0.050s\\
     Sine+PSC & 1/16 & 38.629 & 0.966 & 0.041 & 30.305 & 0.669 & 0.158 
     & 0.204M & 0.050s\\
     \textbf{CHOIR} & \textbf{1/8} & \textbf{41.316} & \textbf{0.992} & \textbf{0.014}  & \textbf{32.605} & \textbf{0.847} & \textbf{0.130} 
     & 0.204M & 0.064s\\
    \bottomrule
  \end{tabular}
\end{table}

\begin{figure}[htbp]
  \centering
  \begin{subfigure}[b]{0.45\linewidth}
    \centering
    \includegraphics[width=\linewidth]{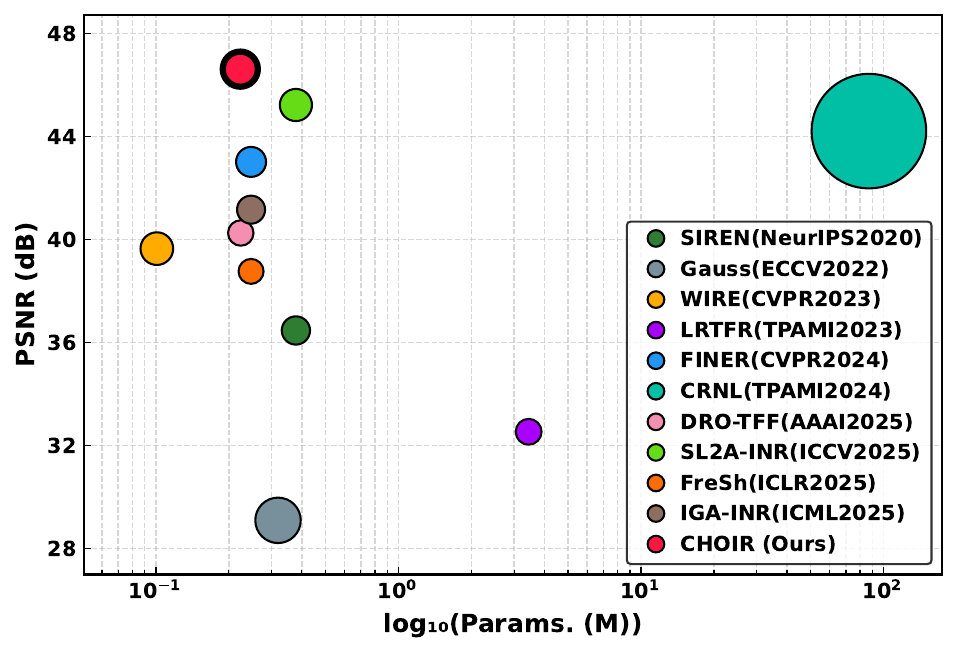}
    \caption{}
    \label{fig:performance}
  \end{subfigure}
  \hfill  
  \begin{subfigure}[b]{0.53\linewidth}
    \centering
    \includegraphics[width=\linewidth]{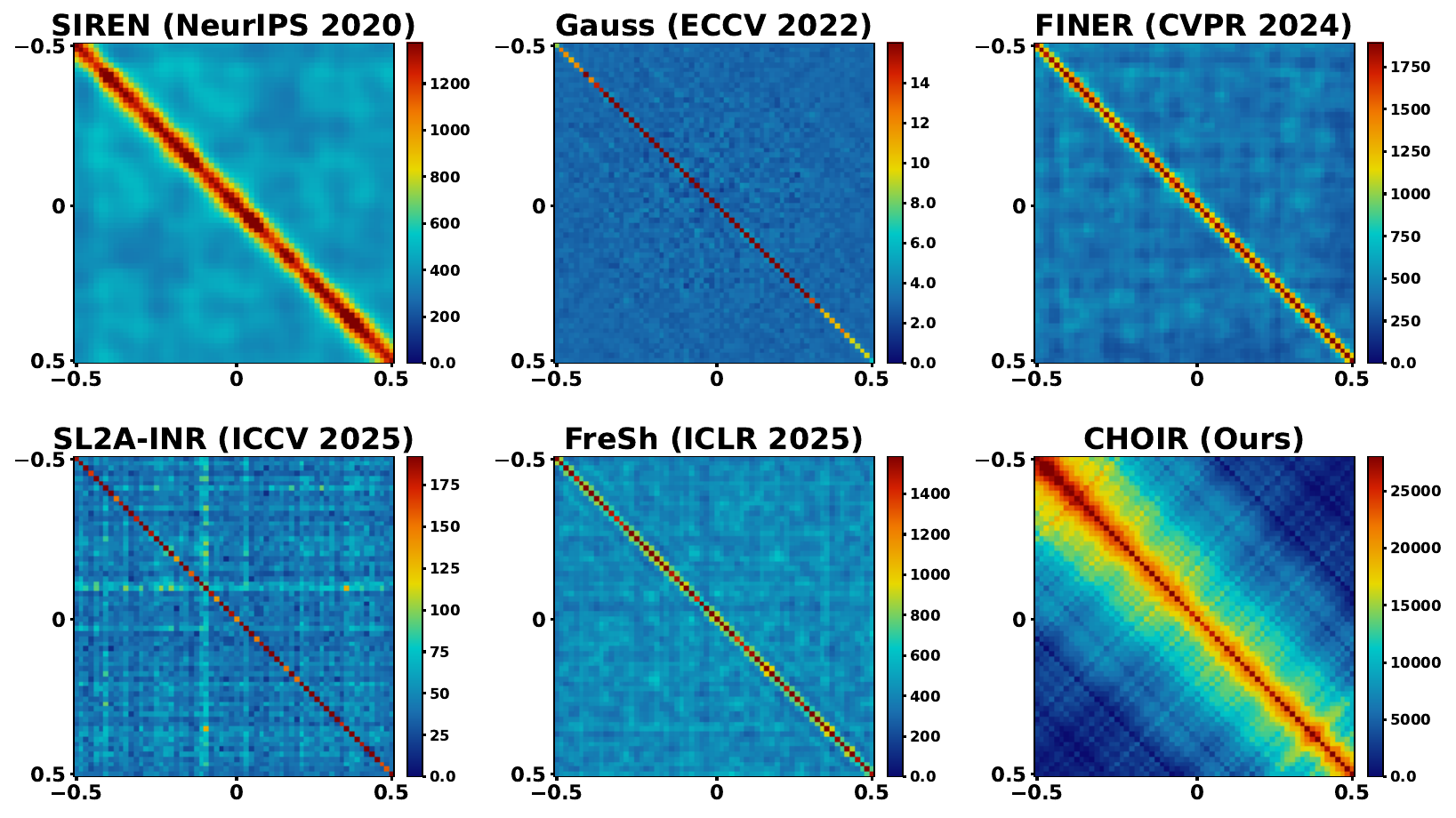}
    \caption{}
    \label{fig:ntk}
  \end{subfigure}
  \caption{(a) Comparison of various INR-based methods on HSI WDC dataset under data missing completion (random missing, OR=0.10). The size of each circle represents the runtime. (b) Visualization of NTK matrices at initialization for various INR methods.}  
  \label{fig:combined}
\end{figure}

\noindent \textbf{Comparative Analysis of Method Complexity.} As shown in Figure~\ref{fig:combined} (a), we compare various INR-based methods on the HSI WDC dataset under random missing completion (OR=0.10) in terms of PSNR, model parameters (logarithmic scale), and runtime. The horizontal axis represents ($\log_{10}$) (Params. (M)), the vertical axis denotes PSNR (dB), and the size of each circle indicates runtime. The proposed CHOIR method achieves the highest PSNR with relatively few parameters and short runtime. In contrast, although CRNL~\cite{luo2024revisiting} attains competitive PSNR, it requires an excessively large number of parameters and long runtime. Lighter methods such as SIREN~\cite{sitzmann2020implicit}, Gauss~\cite{ramasinghe2022beyond}, and WIRE~\cite{saragadam2023wire} remain in the low-parameter regime but yield lower PSNR performance. Overall, CHOIR offers a better trade-off among reconstruction quality, model parameter count, and computational efficiency.

\noindent \textbf{Neural Tangent Kernel (NTK) Analysis.} The Neural Tangent Kernel (NTK) characterizes the gradient similarity between different input coordinates. Its structure at initialization reflects the inductive bias of an untrained network as well as its ability to perform decoupled updates. The degree of diagonal dominance in the NTK indicates the independence of gradient updates across spatial coordinates. A stronger diagonal with weaker off-diagonal elements (corresponding to a darker blue background in the figure) implies that gradient updates are more spatially decoupled, thereby enabling stronger fitting capacity, particularly for capturing high-frequency details. As shown in Fig.~\ref{fig:combined} (b), CHOIR exhibits the strongest diagonal dominance, accompanied by the darkest blue background. This observation theoretically validates that CHOIR possesses a more favorable inductive bias and superior high-frequency representation capability, which is consistent with the conclusions presented in Sec.~\ref{sec:representation}. A more detailed introduction to the NTK matrix is provided in the supplementary materials.



\section{Conclusion}

We propose CHOIR to fundamentally address the optimization instability that has long plagued periodic INRs, which arises from their nested function composition architecture. With our novel Coordinated Harmonic Superposition (CHS) and Perceptual Spectrum Calibration (PSC) mechanisms, CHOIR enables stable training of deep periodic INRs and efficient data-adaptive spectrum learning. Comprehensive experiments demonstrate that CHOIR consistently outperforms state-of-the-art methods and sets new state-of-the-art benchmarks across a variety of multidimensional data recovery tasks. For future work, we aim to extend our experimental evaluations to unstructured or irregularly sampled data, such as point clouds, event-camera streams, and spatial transcriptomics, to further validate the generalization and effectiveness of our model.



\section*{Acknowledgements}
This work was supported by the National Natural Science Foundation of China (Grant Nos. 62272313 and 62372302).

%
%
\bibliographystyle{splncs04}
\bibliography{main}

\end{document}